\documentclass[10pt,leqno]{amsart}
\usepackage{graphicx}
\usepackage{indentfirst,csquotes}

\usepackage{amssymb,amsthm,amsmath}
\usepackage{xcolor,paralist,hyperref,fancyhdr,etoolbox}

\hypersetup{ colorlinks=true, linkcolor=blue, filecolor=black, urlcolor=blue }

\usepackage{float}
\usepackage{booktabs}
\begin{document}
\title{VIS-DICT: A Visual Dictionary for Missing Modality Imputation in Social Network Depression Detection} 

\author{
Hamed Marvi$^{1}$,
Mohammad Mehdi Keikha$^{2}$,
Abolfazl Nadi$^{1,*}$
}

\maketitle

\begin{center}
$^{1}$Department of Computer Science, School of Mathematics, Statistics and Computer Science, College of Science, University of Tehran, Tehran, Iran\\[0.5em]
$^{2}$Department of Computer Science, University of Sistan and Baluchestan, Zahedan, Iran\\[1em]

\texttt{hamed.marvi@ut.ac.ir},
\texttt{keikha@cs.usb.ac.ir},
\texttt{a.nadi@ut.ac.ir}
\end{center}

\let\thefootnote\relax
\footnotetext{$^*$Corresponding author: \texttt{a.nadi@ut.ac.ir}}

\begin{abstract}
Tracking social media posts can help spot early signs of depression. Recent studies show that combining text and images works better for detecting depression than using text alone. However, many social media posts do not have images, which makes it hard to use multimodal models. Most existing methods fill in missing images using retrieval or generative models that need extra training. In this paper, we introduce Vis-Dict, a dictionary-based method that builds missing visual features by linking words to average image vectors from complete training posts. These estimated visual features are then combined with text to track changes in user behavior over time. We tested Vis-Dict on a social media dataset using user timelines of up to 512 posts and compared it with other missing-data methods. The results show that Vis-Dict performs on par with generative networks, reaching an $F_1$-score of 0.9454 and an ROC-AUC of 0.9890. Most importantly, Vis-Dict achieves this strong performance with zero trainable parameters for image generation. These findings show that directly connecting words to visual features is an effective and practical way to handle missing images in depression detection systems.
\end{abstract}

\bigskip

\section{Introduction}

%1
Depression is a major health concern that impacts millions of people globally. It often damages a person’s mental well-being and disrupts their ability to manage daily tasks and responsibilities. Because the signs often appear gradually many people fail to notice them for a long time. Leaving this condition unmanaged is dangerous because it can progress and eventually lead to self-harm or suicide \cite{WHO}. Proactive screening is an important part of mental health care. By identifying these issues early we can prevent the progression of the disorder and help people receive the support they need to recover \cite{Halfin2007}.

%2
Depression is usually diagnosed through clinical interviews and standardized questionnaires by mental health psychologists. During these assessments, psychologists consider the medical history, current symptoms, and observed behavior of individuals with depressive disorder. Standardized questionnaires also help measure the severity of symptoms and how they affect daily life. Although these methods are the main approaches for screening and diagnosing depression, they depend on individuals with depressive disorders being willing to describe their thoughts, feelings, and experiences with honesty \cite{Corrigan2004}.

%3, 4, 5
Early detection of depression remains difficult because traditional clinical assessment depends on individuals recognizing their symptoms and seeking professional help. Many people delay seeking treatment because early symptoms can be mistaken for temporary stress or fatigue. Also some may be worried about stigma and that could discourage them from visiting a clinic or openly discussing their condition. Because of these, depression is often diagnosed only after symptoms have become more severe and this limits the effectiveness of traditional assessment methods for early detection \cite{Corrigan2004,Halfin2007}.

%6
Because traditional clinic visits have so many limits, scientists are now looking at different ways to track health. Instead of waiting for individuals with depressive disorders to speak up, experts are turning to the study of daily behavior. Depression often changes how a person talks, how they express their feelings, and how they interact with others. It can also change their daily habits or their activity levels. These small shifts in behavior usually happen slowly over a long time. Because a psychologist cannot see these changes during a short meeting, analyzing long-term behavioral data is a much better way to spot the first signs of trouble\cite{ReeceEtAl2017, Onnela2016}.

%7
During clinical assessments, some individuals may be reluctant to openly discuss their depressive symptoms because they are concerned about how this information will be perceived or used \cite{Kristofersdottir2026}. This is why social media platforms are such an interesting area for research. These websites offer a space where users share their daily experiences and feelings without the formal constraints of a clinic. While a post on a social media platform does not capture a person's entire mental health situation, it does provide data that feels much more natural than a clinical interview. Because users post content regularly in their daily lives, researchers can track these digital signals over time. For this reason, platforms like Twitter(X) have become widely used for developing and evaluating depression detection models \cite{DeChoudhury, Coppersmith}.

%8
More usage of social media platforms has made large amounts of user generated data available for depression research. Early studies mainly relied on textual information to identify signs of depression. As social media platforms evolved to include images, audio, and videos, researchers began using multimodal learning to combine information from different data types \cite{Baltrusaitis2019Multimodal}. By using multiple modalities, these models can capture complementary behavioral cues that are not available from text alone. This leads to improved depression detection performance \cite{Zhu2023Sentiment,Haque2025MMFformer}.

%9
Although multimodal models have better results, data from social media platforms contains missing modalities. For example a user may share a text only post, an image without a caption, or a video without accompanying text. So the available modalities can vary across posts even within the same user's timeline. Multimodal models assume that all required modalities are available for each sample. When one or more modalities are missing, multimodal fusion becomes challenging, and the effectiveness of the model depends on how the missing modalities are handled \cite{Ma,Zhang2024,RenjieWu2025}.

%One common way to fix this is to simply delete any posts that are missing a part, but that forces us to throw away a huge portion of the user’s history and valuable behavioral clues. Another approach is to fill in the empty space with zeros or other fixed values, but these do not add any real meaning to the analysis. Because of this, we need better ways to handle incomplete data. Effective depression detection models must be able to work with these partial posts without ignoring the useful information that is still available in the other modalities.

%10
Recent research has shown that depression can be reflected in subtle language patterns on social media platforms, such as pronoun usage and word choice \cite{DeChoudhury, Edwards2017}. However, existing missing modality methods are designed as general-purpose recovery techniques that estimate missing modality representations using learned reconstruction models, without explicitly maintaining the correspondence between textual words and the recovered visual representation \cite{Ma, Zhang2024, RenjieWu2025}. This motivates the development of a dictionary-based approach that directly associates specific words with visual embeddings.

%11
To address this issue, we propose Vis-Dict for handling missing visual modalities in multimodal depression detection. We evaluate the proposed method on a depression detection dataset from a social media platform \cite{Shen2017MDDL} and compare it with several existing missing modality methods. The results demonstrate that Vis-Dict is an effective approach for this task. The rest of this paper is organized as follows. Section 2 reviews related work. Section 3 presents the proposed method. Section 4 describes the experiments and discusses the results. Section 5 concludes the paper.

%__________________________________________

\section{Related Work}

This section reviews previous work on multimodal depression detection and methods for handling missing modalities.

\subsection{Multimodal Depression Detection}

Early machine learning studies for depression detection mainly relied on textual information from social media platforms such as Twitter \cite{DeChoudhury,Coppersmith}. These studies showed that language can provide useful information about depressive symptoms. For example, people with depression often use more first-person singular pronouns and more negative words than other users \cite{Edwards2017}. But relying only on text may not capture all of the information shared on social media platforms \cite{Kelley2022}.

As social media platforms began to include images, videos, and other modalities, researchers started combining different data modalities for depression detection \cite{Baltrusaitis2019Multimodal,Guo}. Images can provide information that is not available from text alone. For example studies on Instagram found that visual features like color, brightness, and photo composition are associated with depression \cite{ReeceDanforth2017}. So combining text and visual information provides a better representation of user behavior.

Early multimodal models combined features from different modalities using simple fusion methods \cite{Fang2023,Wang2022}. More recent studies have introduced transformer-based and attention-based models to capture the relationship between text and visual information \cite{Bucur2023,Saeed2025}. These methods have achieved better performance than single modality models and have become the main direction of research in automated depression detection \cite{Gui2019,Xu2020,Pu2024,Tang2025}.

\subsection{The Challenge of Missing Data}

Multimodal learning assumes that information from all modalities is available during both training and inference. But in real world data this assumption is not always true. Social media posts often have missing image modality, missing text modality, or other incomplete modalities. When one or more modalities are unavailable, many multimodal models cannot directly process the incomplete input because they are designed to operate on complete multimodal data \cite{Ma,RenjieWu2025}. Because of that, handling missing modalities is an important research problem in multimodal learning.

One of the simplest ways to handle missing modalities is to replace the missing information with predefined values such as zeros, random vectors, or learned embeddings. These approaches are computationally efficient and are commonly used as baseline strategies for missing modality learning \cite{RenjieWu2025}. Earlier work on multimodal data imputation also explored simple statistical techniques including zero filling, mean, and median imputation to estimate missing values rather than discarding incomplete samples \cite{Campos2015}. These methods allow models to process incomplete data but they do not recover the semantic information contained in the missing modality. As a result, the resulting representations only have limited information and often fail to capture the complex relationships between modalities.

Retrieval and composition based methods recover missing modalities by using information from other samples instead of generating them. One example is Modal-Mixup \cite{YanwuYang2024} which creates new complete samples by combining the available modalities of two incomplete samples from the same class. Another example is MISSRAG \cite{Pipoli2025MISSRAG} which retrieves the most similar samples based on the available modalities and uses the retrieved modality to fill in the missing one. Although these methods are simpler than generative approaches, their performance depends on finding similar samples. If suitable samples are not available or contain unrelated information, the recovered modality may be inaccurate.

Generation-based methods reconstruct the missing modality by generating new data instead of retrieving existing samples. One example is the Multimodal Autoencoder (MMAE) proposed by Jaques et al. \cite{Jaques2017}. MMAE is trained with complete data while randomly hiding one or more modalities during training, forcing the model to predict the missing information from the remaining modalities. This allows the model to learn relationships between different modalities and reconstruct missing features during inference. Also diffusion based methods have shown good performance for missing modality recovery. For example AMM-Diff \cite{Kebaili2025} learns a shared representation from the available MRI modalities and uses a diffusion model to generate the missing modality while also reconstructing the available ones. Although generation based methods often produce more realistic and flexible reconstructions than retrieval-based approaches, they are generally more computationally expensive and require larger amounts of training data to learn accurate data distributions.

Representation recovery methods estimate the latent representation (embedding) of a missing modality instead of reconstructing the original data. One example is SMIL \cite{Ma} which predicts the missing modality's embedding from the available modality and combines it with the available embedding for classification. To improve robustness when only a small amount of complete multimodal data is available, SMIL models uncertainty during feature reconstruction and optimizes the network using a Bayesian meta learning framework. Another example is TRML \cite{Zhao2025TRML} which also predicts the missing modality's embedding from the available one, but further applies contrastive learning to align the estimated embedding with the real embedding of the same sample. These methods are generally more efficient than reconstructing the original modality because they only recover compact feature representations. 

Overall, existing methods have improved missing modality learning through retrieval, generation, and representation recovery. But these methods are designed as general missing modality solutions rather than for depression detection which leaves room for task specific approaches.

%__________________________________________

\section{Proposed Method}
This section describes our framework for detecting depression on social media by combining text and visual data, with a focus on handling missing images. Our goal is to spot signs of depression by looking at a user's posting history over time. To do this, we tackle two main challenges: combining different types of data (text and images) and dealing with the fact that most posts do not include images. Simple fixes, such as replacing missing images with zeros, add useless noise to the model. Because of this, we use context-aware methods to estimate missing visual features without losing important information.

\begin{figure*}[!t]
    \centering
    \includegraphics[width=\textwidth]{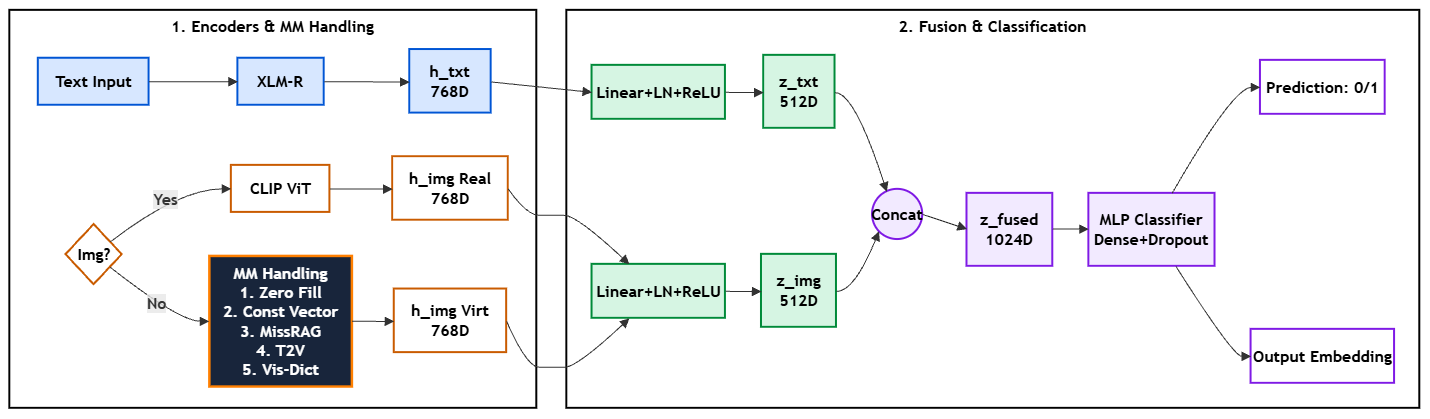}
    \caption{Overall architecture of the tweet-level pipeline. The system extracts text and image representations ($h_{\text{txt}}, h_{\text{img}} \in \mathbb{R}^{768}$), estimates missing visual embeddings using five alternative imputation methods when an image is absent, projects both representations to $\mathbb{R}^{512}$, and concatenates them into a unified vector ($z_{\text{fused}} \in \mathbb{R}^{1024}$) for single-tweet classification and user-level sequence modeling.}
    \label{fig:overall_pipeline}
\end{figure*}

Our framework works in two stages: the tweet level and the user level. As shown in Figure~\ref{fig:overall_pipeline}, at the tweet level, we use XLM-RoBERTa \cite{Conneau2020XLMR} to extract features from text and CLIP ViT \cite{Radford2021CLIP} to extract features from images. We then test five different strategies to fill in the missing image representations. 

Next, we move to the user level to track how a person's behavior changes over time. We compare basic methods like averaging scores with a Sequential Transformer that reads tweets in their exact posting order. This helps the model tell the difference between short-term mood swings and long-term depression. The following subsections explain our feature extraction, the five missing-image methods, and our user-level detection model.

\subsection{Feature Extraction and Modality Encoders}

To allow our model to process both text and images, we convert raw social media posts into dense number vectors. We employ state-of-the-art pretrained encoders. to extract the core meaning from both the text and the images.

\begin{figure}[htbp]
\centering
\includegraphics[width=0.85\textwidth]{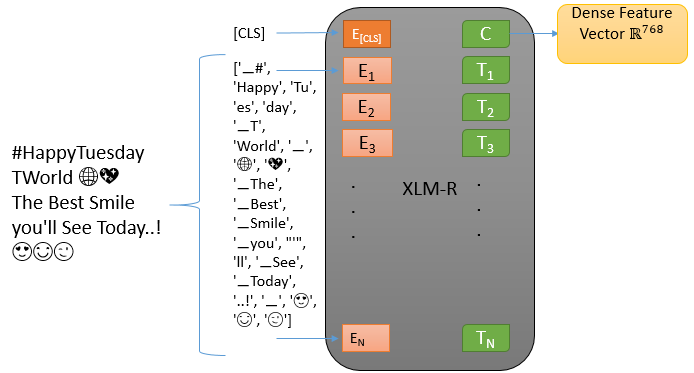}
\caption{Architecture of the textual feature extraction pipeline using XLM-RoBERTa \cite{Conneau2020XLMR} to generate a 768-dimensional dense vector.}
\label{fig:singletext}
\end{figure}

For the text, we use the pre trained multilingual XLM-RoBERTa model \cite{Conneau2020XLMR}. We chose this model because it is very good at understanding different languages and the informal slang often used on social media platforms. As shown in Figure \ref{fig:singletext}, the raw text is broken into tokens, and a special classification token (\textit{CLS}) is added at the beginning. The model turns these tokens into a single vector that captures the meaning of the whole text:
\begin{equation}
\mathbf{h}_{txt} = \text{Encoder}_{XLM\text{-}R}(\text{Text}) \in \mathbb{R}^{768}
\end{equation}

\begin{figure}[htbp]
\centering
\includegraphics[width=0.9\textwidth]{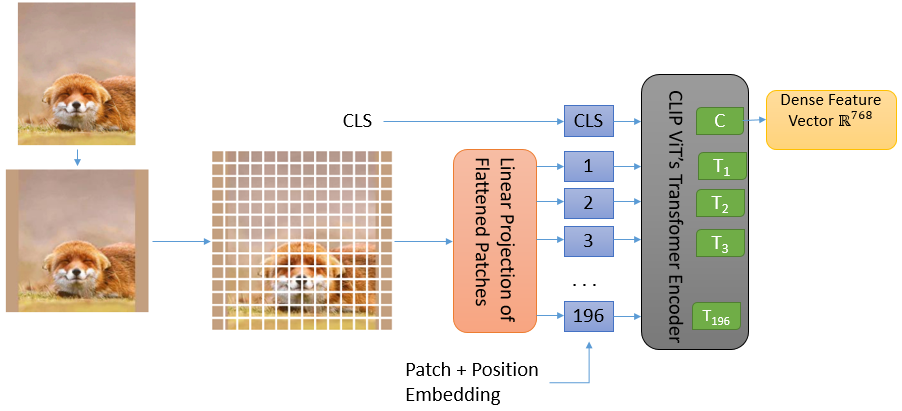}
\caption{Architecture of the visual feature extraction pipeline using CLIP ViT \cite{Radford2021CLIP}, illustrating the custom resizing and average-color padding strategy.}
\label{fig:singleimage}
\end{figure}

For the images, we extract features using the CLIP Vision Transformer (ViT) \cite{Radford2021CLIP}. To fit the required input size of the model without stretching or distorting the images, we carefully resize them. We resize the longest side of the image to 224 pixels and fill the empty space with the average color of the image to create a standard $224 \times 224$ square. As shown in Figure \ref{fig:singleimage}, this resized image is split into a grid of patches. The model processes these patches to create a single vector that represents the whole image:
\begin{equation}
\mathbf{h}_{img\_raw} = \text{Encoder}_{ViT}(\text{Image}) \in \mathbb{R}^{768}
\end{equation}

We evaluate two different training strategies for our image model on complete samples (posts containing images):
\begin{itemize}
    \item \textbf{Unjoint Training (Independent Fine-Tuning):} We fine-tune the entire CLIP Vision Transformer encoder along with a linear classification head using only the classification cross-entropy loss ($\mathcal{L}_{cls}$). In this setting, the visual model learns depression-related features independently, without any interaction or alignment with the textual representations.
    \item \textbf{Joint Training (Cross-Modal Visual Alignment):} To map visual features into the semantic space of text, we align the visual representations with fixed textual embeddings extracted from XLM-RoBERTa. We train the CLIP Vision Transformer, a linear projection layer, and the classifier together using a combined multi-task loss function:
    \begin{equation}
    \mathcal{L}_{Joint} = \mathcal{L}_{cls} + \lambda \cdot \mathcal{L}_{align}
    \end{equation}
    where $\lambda = 0.5$ balances the two objectives. The alignment loss minimizes the cosine distance between the projected visual embedding ($\mathbf{h}_{img}$) and the corresponding text anchor ($\mathbf{h}_{txt}$):
    \begin{equation}
    \mathcal{L}_{align} = 1 - \frac{\mathbf{h}_{img} \cdot \mathbf{h}_{txt}}{\|\mathbf{h}_{img}\| \|\mathbf{h}_{txt}\|}
    \end{equation}
\end{itemize}

\subsection{Addressing the Missing Modality Challenge}

In real world data from social media platforms, posts are often incomplete. Users frequently write text without including a picture. This missing data creates a problem for our models. Simple ways to fill in the missing image usually fail because they do not capture the true meaning of the post. 

To solve this problem and avoid losing information, we test five different ways to impute missing representations. These methods range from very simple replacements to context-aware approaches that estimate the missing image based on the text. Using these strategies keeps the input size exactly the same for every post, whether it has a real image or not.

\subsubsection{Baseline Imputation}
We first evaluate two fundamental imputation baselines that serve as the lower bound for handling missing visual modalities:

\begin{enumerate}
    \item \textbf{Zero-Filling:} In this configuration, missing image features are replaced with a null vector, essentially performing an identity mapping of zero valued information into the fusion pipeline:
    \begin{equation}
    \mathbf{h}_{img} = \mathbf{0} \in \mathbb{R}^{768}
    \end{equation}
    While computationally trivial, this assumes that the absence of a visual modality conveys zero information. In practice, this forces the subsequent fusion layers to ingest a non informative feature distribution, which often injects noise into the representation space and disrupts the convergence stability of the downstream classifier.

    \item \textbf{Constant Learned Vector:} To alleviate the sparsity issues inherent in zero-filling, we introduce a globally trainable parameter $\mathbf{v}_{\text{miss}} \in \mathbb{R}^{768}$. This vector acts as a learnable placeholder that is optimized via backpropagation alongside the primary network weights:
    \begin{equation}
    \mathbf{h}_{img} = \mathbf{v}_{\text{miss}}
    \end{equation}
    Unlike the zero-filling baseline, this approach allows the network to adaptively discover the most representative global visual prior for text only posts. By treating the missing modality as a learnable entity, the model converges toward a more optimal latent representation than the null hypothesis provided by zero-filling.
\end{enumerate}

\subsubsection{Context Aware Baselines}
\begin{figure}[htbp]
    \centering
    \includegraphics[width=0.7\textwidth]{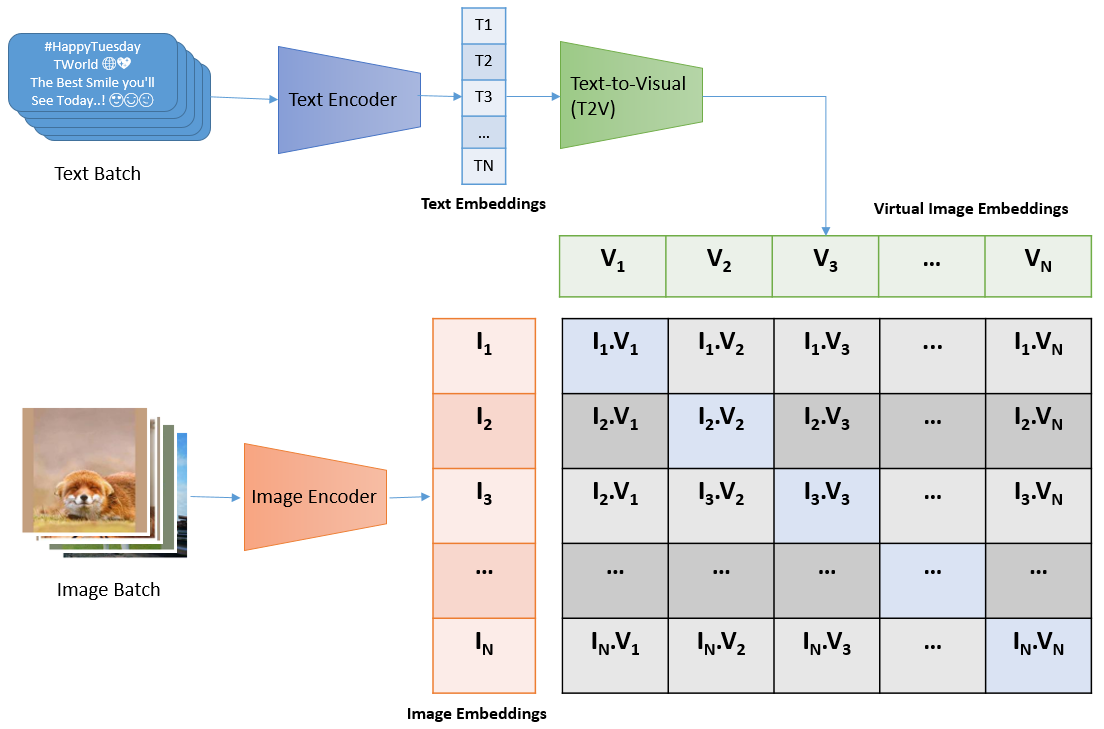}
    \caption{Architecture of the Text-to-Visual (T2V) translation module.}
    \label{fig:t2v_noloss}
\end{figure}

The simple baseline methods are stable, but they ignore the text of the post. To see how well models can estimate missing images based on the text, we test two advanced baselines:

\begin{enumerate}
    \item \textbf{Feature Level MissRAG (Hybrid):}We adapt the retrieval mechanism described in MissRAG for feature-level visual embedding recovery, , where retrieved visual representations are used to impute missing image embeddings. \cite{Pipoli2025MISSRAG}. This method searches a database of posts that have both text and images ($\mathcal{D}_{complete} \subset \mathcal{D}_{train}$). For a given text vector $\mathbf{h}_{txt} \in \mathbb{R}^{768}$, we measure how similar it is to the posts in the database using cosine similarity. If the similarity score is $\ge 0.5$, we average the image vectors of the top 5 matches to fill in the missing image. If the score is lower, the model defaults to the learned placeholder $\mathbf{v}_{\text{miss}}$. This ensures we only use retrieved images when we find a strong match.

    \item \textbf{Text-to-Visual (T2V) Embedding Translation:} Following the text-to-visual mapping component introduced in TRML \cite{Zhao2025TRML}, we implement a neural network that translates text embeddings directly into visual embeddings (Figure \ref{fig:t2v_noloss}). Unlike the original method, which additionally employs contrastive learning for cross-modal alignment, we only optimize the generated visual embedding to match the corresponding ground-truth visual embedding using reconstruction-based objectives. We train this generator $\hat{\mathbf{v}} = f(\mathbf{h}_{txt})$ using a loss function that reduces both the distance and the angle difference between the generated vector and the real one:
    \begin{equation}
    \mathcal{L}_{T2V} = \frac{1}{d} \sum_{k=1}^{d} (\hat{\mathbf{v}}_k - \mathbf{v}_k)^2 + \left( 1 - \frac{\hat{\mathbf{v}} \cdot \mathbf{v}}{\|\hat{\mathbf{v}}\| \|\mathbf{v}\|} \right)
    \end{equation}
    where $\mathbf{v}$ is the real image vector. This forces the generated image vector to closely match the true size and direction of the original image.
\end{enumerate}

\subsubsection{Proposed TF-IDF Visual Dictionary - Vis-Dict}

\begin{figure}[htbp]
\centering
\includegraphics[width=0.95\linewidth]{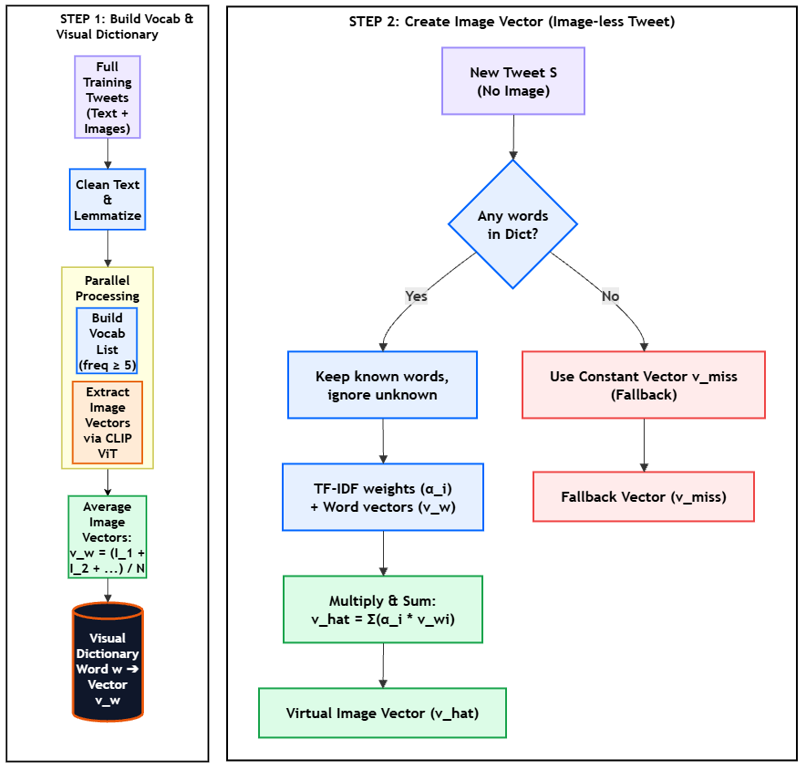}
\caption{Overview of the Vis-Dict framework: dictionary construction from complete multimodal posts (Step 1) and feature imputation for posts without images (Step 2).}
\label{fig:visdict}
\end{figure}

To capture word-level visual cues for posts without images, we propose Vis-Dict. The overall process for building the dictionary and estimating missing image features is shown in Figure \ref{fig:visdict}.

\textbf{Dictionary Construction:} We first clean the text by removing URLs and tags, applying lemmatization \cite{Bird2009NLTK}, and removing stop-words. We build the dictionary $\mathcal{V}_{vis}$ using only the training data that has both text and images ($\mathcal{D}_{complete} \subset \mathcal{D}_{train}$). In parallel, we build a vocabulary list of words that appear at least 5 times ($\tau=5$) to filter out rare words, and extract image vectors using the CLIP ViT encoder. For each word $w \in \mathcal{V}_{vis}$, we calculate an average image vector $\mathbf{v}_w \in \mathbb{R}^{d}$:
\begin{equation}
\mathbf{v}_w = \frac{1}{|P_w|} \sum_{I \in P_w} \text{Encoder}_{ViT}(I)
\end{equation}
where $P_w$ is the set of images from posts containing the word $w$ when both text and image modalities are available.

\textbf{Feature Imputation:} When a post $i$ has text words $S_i$ but is missing an image, we create a new image vector $\hat{\mathbf{v}}_{I,i}$. If the post contains words from the dictionary, we take the average image vectors of those words and combine them using normalized TF-IDF weights:
\begin{equation}
\hat{\mathbf{v}}_{I,i} = \sum_{w_j \in S_i \cap \mathcal{V}_{vis}} \alpha_{j} \cdot \mathbf{v}_{w_j}
\end{equation}
The weights $\alpha_j$ are calculated as \cite{Salton1988TFIDF}:
\begin{equation}
\alpha_j = \frac{\text{TF-IDF}(w_j, S_i)}{\sum_{w_k \in S_i \cap \mathcal{V}_{vis}} \text{TF-IDF}(w_k, S_i)}
\end{equation}
If a post contains no words from the visual dictionary, the missing image representation defaults to the constant learned vector $\mathbf{v}_{\text{miss}}$.

Because this method relies on direct word-to-image averages, it does not introduce additional trainable parameters during the imputation step, keeping the estimated visual vector directly tied to the words present in the text.

\subsection{Multimodal Data Fusion and Classification}
To combine the text and image features, we pass them through a few network layers to create a single, shared representation. Let $\mathbf{h}_{txt} \in \mathbb{R}^{768}$ and $\mathbf{h}_{img} \in \mathbb{R}^{768}$ be the extracted text and image vectors. If the image is missing, we replace $\mathbf{h}_{img}$ with the vector generated by one of our fill in strategies.

First, we project both types of data into a smaller 512-dimensional space:
\begin{equation}
\mathbf{z}_{txt} = \text{LN}(\text{ReLU}(\mathbf{W}_{txt}\mathbf{h}_{txt} + \mathbf{b}_{txt}))
\end{equation}
\begin{equation}
\mathbf{z}_{img} = \text{LN}(\text{ReLU}(\mathbf{W}_{img}\mathbf{h}_{img} + \mathbf{b}_{img}))
\end{equation}
where $\mathbf{W}_{txt}, \mathbf{W}_{img} \in \mathbb{R}^{512 \times 768}$ are weight matrices, $\mathbf{b}_{txt}, \mathbf{b}_{img} \in \mathbb{R}^{512}$ are biases, and LN stands for Layer Normalization \cite{Ba2016LN}. This step helps the training process stay stable.

Next, we combine these two smaller vectors into one 1024-dimensional vector:
\begin{equation}
\mathbf{z}_{fused} = [\mathbf{z}_{txt}; \mathbf{z}_{img}] \in \mathbb{R}^{1024}
\end{equation}

Finally, this combined vector goes through a final classification layer to give us our predictions:
\begin{equation}
\mathbf{h}_{dense} = \text{Dropout}(\text{ReLU}(\mathbf{W}_{1}\mathbf{z}_{fused} + \mathbf{b}_{1})) \in \mathbb{R}^{256}
\end{equation}
\begin{equation}
\hat{\mathbf{y}} = \mathbf{W}_{2}\mathbf{h}_{dense} + \mathbf{b}_{2} \in \mathbb{R}^{2}
\end{equation}
where $\mathbf{W}_{1} \in \mathbb{R}^{256 \times 1024}$ and $\mathbf{W}_{2} \in \mathbb{R}^{2 \times 256}$. We use a dropout rate of 0.3 \cite{Srivastava2014Dropout} here to prevent the model from overfitting. This setup ensures the model processes the data exactly the same way, whether the image is real or filled in.

\subsection{User-Level Detection}
After classifying individual tweets, we need to look at the user's entire history to make a final decision. We test two simple aggregation methods and one sequence model to do this:

\begin{itemize}
    \item \textbf{Majority Voting:} Each post is scored independently. A user is classified as depressed if the number of their depressed posts is higher than a specific threshold $\theta$.
    
    \item \textbf{Probability Averaging:} This method takes the model's confidence scores for every post and calculates the average across the user's whole timeline. If this average score is higher than our threshold $\theta$, the user is labeled as depressed.
    
    \item \textbf{Sequential Transformer:} Basic math methods ignore the order of the tweets. To capture how behavioral patterns change over time, we use a Sequential Transformer. Let the timeline of a user’s combined tweet vectors be $\{e_1, e_2, ..., e_T\}$, where $e_t \in \mathbb{R}^{1024}$. We first project each vector into a smaller space using a linear layer: $h_t = W_{proj}e_t \in \mathbb{R}^{256}$.
    
    For our single-modality user-level models (such as pure text timelines or pure image timelines), the inputs are individual 768-dimensional tweet representations instead of fused 1024-dimensional vectors. In these single-modality settings, we skip the multimodal fusion step entirely and project the 768-dimensional single vectors directly into the exact same 256-dimensional space ($h_t \in \mathbb{R}^{256}$) using a matching linear projection layer. This structural design ensures that the internal transformer parameters and dimensions stay completely identical across all single-modality and multimodal experiments for a fair evaluation.

    We process sequences of up to $T=512$ tweets per user. If a user has fewer tweets, we use pre-padding by filling the start of the sequence with zeros. To help the model understand the exact chronological order of the posts, we add learned positional encodings $p_t$ to the sequence:
    \begin{equation}
    X_{user} = \text{Transformer}(\{h_t + p_t\}_{t=1}^{T}) \in \mathbb{R}^{T \times 256}
    \end{equation}
    The Transformer uses self-attention to find long-term patterns in the user's behavior. We then use masked mean pooling to average the vectors of the actual (non-padding) tweets. This final user representation is passed through a classification layer to get the final prediction $\hat{y}_{user}$. This allows us to analyze the timeline as an ongoing story of behavior rather than an unordered collection of posts.
\end{itemize}

The next section will show the experimental results for these three methods.

\section{Experiments and Results}
In this section, we evaluate our proposed framework and how it addresses the missing image problem. First, we describe our dataset and the setup used to train our models. Then, we analyze the results at the tweet level to understand how text and image features work together. Finally, we test our full timeline model at the user level to show how well our different missing-image strategies perform.

\subsection{Dataset and Experimental Setup}
To test our framework, we used a subset of the social media data popularized by Shen et al.~\cite{Shen2017MDDL}, categorizing users into two groups: Normal (Class 0) and Depressed (Class 1). In the original MDDL dataset, Shen et al.~\cite{Shen2017MDDL} established user-level labels using rule-based heuristics so that users were classified as depressed if they posted tweets strictly matching patterns like ``I am diagnosed depression,'' ``I'm diagnosed depression,'' ``I've been diagnosed depression,'' or ``I was diagnosed depression,'' while normal users were identified by a complete absence of the string ``depress'' in their posting history. 

However, because the explicit self-report tweet (the anchor tweet) directly announces the diagnosis, keeping it in the dataset would cause data leakage and artificially inflates evaluation performance. To prevent this leakage, the very first step in our preprocessing was to chronologically sort each user's timeline and remove the final anchor tweet containing the self-report statement from all depressed users before any feature extraction, dictionary construction, or model training.

Because depression labels in the dataset are given to entire users, individual posts inherit the user's overall label for single-tweet tests. In real life, a depressed user often posts everyday, neutral, or happy messages that do not show signs of depression. As a result, single-tweet classification includes unavoidable label noise, while tracking a user's timeline over time helps the model learn real behavioral patterns.

While the original dataset serves as a foundational resource for multimodal research, raw social media data is inherently noisy and subject to content attrition over time. Therefore, we implemented a custom data-processing pipeline to ensure the quality required for our longitudinal timeline model. Specifically, we first retrieved available images associated with the text-based entries. We then filtered for users who contributed at least one post containing both text and images to establish a baseline for cross-modal interaction. To ensure a robust evaluation and prevent class imbalance, we performed a user-level timeline matching strategy. For each user in the Depressed group with $n_i$ tweets, we selected a user from the Normal group with at least $n_i$ tweets. If the selected Normal user had more than $n_i$ tweets, only their most recent $n_i$ tweets were retained to preserve the chronological structure of the user's timeline. As a result, each Depressed user was paired with a Normal user containing exactly the same number of tweets, leading to a balanced dataset in terms of both user count and total tweet volume across the two classes. The matching procedure was performed before the train, validation, and test split to maintain independent user-level partitions.

All preprocessing operations that learn information from the data were performed exclusively on the training split. Specifically, the visual dictionary was constructed only using complete multimodal samples from the training users, and retrieval databases used by MissRAG contained only training samples. User-level splitting (80\% train, 10\% validation, and 10\% test) was performed strictly by user ID before feature extraction to ensure that no tweets from the same user appeared across different splits, eliminating inter-split data leakage.

\subsubsection{Dataset Statistics and Modality Sparsity}

To address class imbalance and prevent model bias toward the majority class, we performed undersampling on the Normal (Class 0) cohort, resulting in a balanced distribution of users and posts. As detailed in Table \ref{tab:dataset_splits}, the final dataset comprises 3,504 unique users, partitioned equally into 1,752 Normal and 1,752 Depressed users, with a balanced tweet volume across both groups. The dataset was divided by user identity into Training (80\%), Validation (10\%), and Testing (10\%) splits. This user-level partitioning ensures that a user's temporal activity remains exclusive to a single split, effectively preventing inter-split data leakage.

Our dataset reflects the modality sparsity prevalent in real-world social media environments. Although users contribute an average of 237 posts, approximately 90.4\% of these entries consist solely of text, lacking any visual modality. This significant imbalance reveals a fundamental limitation in standard fusion models that rely on simple zero-filling strategies. When visual data is absent for 90\% of a timeline, the model’s ability to capture comprehensive behavioral patterns is severely constrained, necessitating the context-aware imputation strategies proposed in this paper.

\begin{table}[htbp]
\centering
\caption{User-level dataset distribution and longitudinal timeline statistics across the experimental splits.}
\label{tab:dataset_splits}
\begin{tabular}{lcccc}
\hline
\textbf{Split} & \textbf{Class 0 (Normal)} & \textbf{Class 1 (Depressed)} & \textbf{Avg Tweets / User} & \textbf{Missing Images} \\
\hline
Training   & 1,401 & 1,401 & 235 & 90.6\% \\
Validation & 175   & 175   & 202 & 88.7\% \\
Testing    & 176   & 176   & 290 & 90.1\% \\
\hline
\textbf{Overall} & \textbf{1,752} & \textbf{1,752} & \textbf{237} & \textbf{90.4\%} \\
\hline
\end{tabular}
\end{table}

\subsubsection{Experimental Setup and Implementation Details}
All models were implemented using PyTorch and trained on an NVIDIA GeForce RTX 4090 GPU with 24GB VRAM. In mental health screening, correctly identifying depressed individuals while keeping false alarms low is the primary goal. Therefore, we use the F1-Score of the Depressed class (Class 1) as our primary evaluation metric. During the validation step, we tested different decision thresholds, and we chose a standard threshold of 0.5 for all final test evaluations to keep results consistent.

Feature Extraction: Text posts were tokenized using \texttt{xlm-roberta-base} with a maximum sequence length of 128 tokens. The single-text model was trained using the AdamW optimizer with a learning rate of $2 \times 10^{-5}$, a weight decay of $0.01$, and a linear learning rate scheduler with a $0.10$ warmup ratio for 2 epochs with a batch size of 64. For images, the CLIP ViT-B/32 model ($d=768$) processed $224 \times 224$ images with average-color padding to keep the original aspect ratios.

Modality Imputation:
\begin{itemize}
    \item Vis-Dict (Ours): Text was cleaned with spaCy via lemmatization, punctuation removal, and the removal of common social media stop-words like ``amp'' and ``rt''. The visual dictionary and TF-IDF weights were created using only full modality training posts with words that appear at least 5 times. Virtual image vectors were calculated using TF-IDF weighted averages.
    \item Learned Vector: Initialized as a trainable 768-dimensional random vector and optimized during training.
    \item MissRAG: Retrieved the top 5 most similar complete posts using cosine similarity and averaged their visual vectors.
    \item T2V Network: A neural network trained to map 768-dimensional text vectors to 768-dimensional visual vectors using Mean Squared Error and Cosine Similarity loss.
\end{itemize}

Multimodal Fusion and User-Level Modeling:
For single posts, text and image vectors ($768\text{D}$) were each projected to $512\text{D}$ using a Linear layer, Layer Normalization, and a ReLU activation. These two vectors were joined together ($1024\text{D}$) and passed through a Dense layer ($256\text{D}$), a ReLU activation, Dropout ($p=0.30$), and a final classification layer. This fusion model was trained with Cross-Entropy loss and AdamW ($\text{LR}=2 \times 10^{-5}$, weight decay $0.01$, batch size 64) for 2 epochs.

For user-level modeling, the Temporal Transformer processed timelines of up to 512 tweets in chronological order with zero-padding on the left. The tweet vectors were projected to a hidden size of $d_{model}=256$ and combined with 512 positional embeddings. The transformer encoder had 2 layers, 8 attention heads, a feed-forward size of 1024, and a dropout rate of $0.20$. We used masked mean pooling to ignore padding before the final classification layer. The model was trained using Binary Cross-Entropy with Logits Loss and AdamW ($\text{LR}=1 \times 10^{-4}$, weight decay $0.01$, batch size 32) for 2 epochs across 5 random seeds. Table~\ref{tab:hyperparams} summarizes the main hyperparameters.

\begin{table}[h]
\centering
\small
\caption{Hyperparameter and architectural specifications across model components.}
\label{tab:hyperparams}
\begin{tabular}{lccccc}
\toprule
Component / Model & Learning Rate & Batch Size & Weight Decay & Dropout & Epochs \\
\midrule
Text Encoder (XLM-R) & $2 \times 10^{-5}$ & 64 & 0.01 & - & 2 \\
T2V Mapping Network & $1 \times 10^{-4}$ & 64 & 0.01 & - & 3 \\
Multimodal Fusion Head & $2 \times 10^{-5}$ & 64 & 0.01 & 0.30 & 2 \\
User Temporal Transformer\textsuperscript{*} & $1 \times 10^{-4}$ & 32 & 0.01 & 0.20 & 2 \\
\bottomrule
\multicolumn{6}{l}{\scriptsize \textsuperscript{*}Transformer details: $d_{model}=256$, Layers=2, Attention Heads=8, Max Sequence Length=512.}
\end{tabular}
\end{table}

\subsection{Tweet-Level Evaluation}

Before we look at user behavior over time, we need to test our model on single tweets. In this section, we first test the text and image models separately to see how well they work on their own. Next, we test our combined model to see how it handles missing images. Finally, we look closely at our estimated images to make sure they capture real signs of depression.

\subsubsection{Single-Modality Baselines}

To understand the baseline capability of each data source, we evaluated the text and image models separately at the tweet level. Table \ref{tab:single_modal_baselines} shows the classification performance for both modalities.

The single-modality text model evaluated on all test tweets achieved an Accuracy of 0.7633 and an ROC-AUC of 0.8578 (with a Depressed F1-score of 0.7598). This shows that textual content provides a strong and stable baseline signal for identifying depression indicators in isolated posts.

For the visual branch, we evaluated the CLIP Vision Transformer on posts where genuine images were present. When fine-tuned independently using task cross-entropy (Unjoint), the image model achieved an Accuracy of $0.8676 \pm 0.0165$, an ROC-AUC of $0.9361 \pm 0.0048$, and an F1-score of $0.7911$. However, when trained using the multi-task objective to align with text anchors (Joint), the standalone visual performance decreased to an Accuracy of $0.7937 \pm 0.0280$, an ROC-AUC of $0.8707 \pm 0.0041$, and an F1-score of $0.7206$. This performance drop indicates that constraining the visual space using cosine alignment reduces its standalone classification ability compared to fine-tuning on task labels alone.

\begin{table}[htbp]
\centering
\caption{Performance of single-modality baselines at the tweet level.}
\label{tab:single_modal_baselines}
\small
\begin{tabular}{llccc}
\hline
\textbf{Modality} & \textbf{Evaluated Samples} & \textbf{Accuracy} & \textbf{ROC-AUC} & \textbf{Depressed F1} \\
\hline
Text-Only & All Samples & 0.7633 & 0.8578 & 0.7598 \\
\hline
Image-Only (Unjoint) & Full Modality (Real Images) & $0.8676 \pm 0.0165$ & $0.9361 \pm 0.0048$ & 0.7911 \\
Image-Only (Joint)   & Full Modality (Real Images) & $0.7937 \pm 0.0280$ & $0.8707 \pm 0.0041$ & 0.7206 \\
\hline
\end{tabular}
\end{table}

\subsubsection{Multimodal Classification with Missing Images}

To evaluate how different image imputation methods perform on single posts, we tested all five methods using both unjoint and joint training methods. We evaluated the models on the complete test set, as well as separately on posts with and without images. Table \ref{tab:multimodal_missing} shows the tweet-level classification results, reported as Depressed $F_1$ score and ROC-AUC.

The results show two important patterns in single-tweet classification. First is that every model performs much better on posts that contain real images ($F_1 \approx 0.81\text{--}0.85$, $\text{ROC-AUC} \approx 0.90\text{--}0.93$), which shows that genuine visual content provides useful information. Second, for single tweets, text remains the primary signal and  imputation methods achieve scores ($F_1 \approx 0.74\text{--}0.75$, $\text{ROC-AUC} \approx 0.84\text{--}0.85$) that are close to simple zero-filling and learned-vector baselines.

The single-tweet evaluation reveals a limitation of the dictionary approach. Vis-Dict relies on word-level visual associations and when isolated tweets have sparse or neutral vocabulary, the TF-IDF weighted combination gravitates toward an uninformative, neutral centroid. In contrast, parametric models like T2V learn a continuous generative mapping from dense text features to the visual space. 

However, this limitation is lessened at the user level, where aggregating visual tokens across timelines of up to 512 posts filters out single-post noise.

\begin{table}[htbp]
\centering
\caption{Tweet-level multimodal classification results reported as Depressed $F_1$ and (ROC-AUC) across data subsets.}
\label{tab:multimodal_missing}
\small
\begin{tabular}{llccc}
\hline
\textbf{Imputation Strategy} & \textbf{Training} & \textbf{All Samples} & \textbf{Full Modality} & \textbf{Missing Modality} \\
\hline
Zero Placeholder & Unjoint & 0.7624 (0.8562) & 0.8556 (0.9370) & 0.7477 (0.8474) \\
                 & Joint   & 0.7596 (0.8540) & 0.8166 (0.9050) & 0.7477 (0.8474) \\
\hline
Constant Learnt  & Unjoint & 0.7589 (0.8552) & 0.8556 (0.9370) & 0.7514 (0.8475) \\
                 & Joint   & 0.7576 (0.8518) & 0.8166 (0.9050) & 0.7514 (0.8475) \\
\hline
MissRAG          & Unjoint & 0.7593 (0.8587) & 0.8556 (0.9370) & 0.7470 (0.8472) \\
                 & Joint   & 0.7577 (0.8559) & 0.8166 (0.9050) & 0.7470 (0.8474) \\
\hline
T2V Network      & Joint   & 0.7585 (0.8531) & 0.8166 (0.9050) & 0.7465 (0.8472) \\
\hline
Vis-Dict (Ours)  & Unjoint & 0.7669 (0.8602) & 0.8556 (0.9370) & 0.7467 (0.8489) \\
                 & Joint   & 0.7644 (0.8571) & 0.8166 (0.9050) & 0.7475 (0.8484) \\
\hline
\end{tabular}
\end{table}

\subsubsection{Evaluation of Imputed Visual Representations and Mode Collapse}

To test whether the imputed visual features capture meaningful representations rather than collapsing into trivial constant vectors, we evaluated feature diversity across missing-image samples. Following the variance criterion from the VICReg framework \cite{Bardes2022VICReg}, we measured the average dimensional variance across all $d = 768$ feature dimensions:
\begin{equation}
v(Z) = \frac{1}{d} \sum_{j=1}^{d} \text{Var}(z_{:, j})
\end{equation}
where $\text{Var}(z_{:, j})$ is the variance of the $j$-th feature dimension across the evaluated dataset. A higher variance indicates that the model produces diverse embeddings for different inputs, while a near-zero variance indicates mode collapse.

Table \ref{tab:variance_eval} reports the dimensional variance across missing-image samples compared to the ground-truth visual baseline.

\begin{table}[htbp]
\centering
\caption{Average dimensional variance of visual embeddings on missing-image samples.}
\label{tab:variance_eval}
\small
\begin{tabular}{llc}
\hline
\textbf{Imputation Strategy} & \textbf{Training Paradigm} & \textbf{Dimensional Variance} \\
\hline
Real Image (Oracle Baseline) & --- & 0.6842 \\
MissRAG                      & Joint & 0.4835 \\
T2V Network                  & Joint & 0.4288 \\
MissRAG                      & Unjoint & 0.2252 \\
Vis-Dict (Ours)              & Joint & 0.1387 \\
Vis-Dict (Ours)              & Unjoint & 0.0374 \\
\hline
\end{tabular}
\end{table}

The dimensional variance results in Table \ref{tab:variance_eval} show clear differences across the imputation methods. Deep learning and retrieval baselines, such as the neural T2V network ($0.4288$) and retrieval-based MissRAG ($0.4835$ in joint training), have higher feature variance closer to real images ($0.6842$). In addition, the generated T2V embeddings achieve an average cosine similarity of $0.8773$ with the corresponding text embeddings, confirming an effective continuous cross-modal mapping.

In comparison, the proposed Vis-Dict shows lower dimensional variance, measuring $0.0374$ in the unjoint setting and $0.1387$ in the joint setting. Because Vis-Dict computes representations by taking a TF-IDF weighted average of word-level vectors, short posts that lack strong emotional or domain-specific vocabulary tend to pull the estimated visual feature toward a neutral centroid. 

However, this low single-post variance acts as a regularized prior rather than a failure of the method. While individual imputed posts occupy a compact subspace, aggregating these lightweight visual tokens across a user's longitudinal timeline ($T=512$) cancels out post-level noise and captures consistent behavioral signals.

\subsubsection{Modality Agreement and Error Analysis}

To analyze how the textual and visual branches interact during fusion, we evaluated the agreement between unimodal predictions and the final multimodal output at the tweet level using the Unjoint Vis-Dict configuration. For each branch, predictions were generated through its dedicated classification head using a standard decision threshold of 0.5.

Table \ref{tab:modality_interaction} provides a breakdown of prediction agreements and error patterns across data subsets.

\begin{table}[htbp]
\centering
\caption{Tweet-level modality agreement and error breakdown for the Unjoint Vis-Dict model across data subsets.}
\label{tab:modality_interaction}
\small
\begin{tabular}{lccccc}
\hline
\textbf{Prediction Pattern} & \textbf{Class 0} & \textbf{Class 1} & \textbf{Real Images} & \textbf{Missing Images} & \textbf{Total} \\
\hline
Text Correct, Image Incorrect $\rightarrow$ Correct & 16,770 & 5,267 & 1,030 & 21,007 & 22,037 \\
Image Correct, Text Incorrect $\rightarrow$ Correct & 350    & 1,117 & 608   & 859    & 1,467  \\
Text Incorrect Misleads Fusion                    & 2,309  & 8,314 & 214   & 10,409 & 10,623 \\
Image Incorrect Misleads Fusion                   & 843    & 235   & 487   & 591    & 1,078  \\
Both Incorrect $\rightarrow$ Correct Fusion       & 123    & 21    & 22    & 122    & 144    \\
Both Correct $\rightarrow$ Incorrect Fusion       & 12     & 91    & 10    & 93     & 103    \\
\hline
\end{tabular}
\end{table}

The breakdown demonstrates that text serves as the dominant modality across the dataset. In 22,037 instances, a correct textual prediction preserved the accurate final classification despite an incorrect visual prediction. In 1,467 cases, the visual representation provided informative cues that allowed the network to produce a correct prediction when the text branch alone was inaccurate. Additionally, in 144 cases, combining features from both branches produced a correct output despite individual unimodal misses near the decision boundary.

Conversely, conflicting signals between branches also accounted for classification errors. Incorrect text representations misled the fused prediction in 10,623 instances, whereas inaccurate visual features caused misclassifications in 1,078 cases. Finally, 103 samples resulted in incorrect fused outputs despite correct individual unimodal classifications. These results indicate that while estimated visual features can reinforce classification when textual signals are ambiguous, disagreements between branches remain the primary contributor to single-tweet prediction errors.

\subsection{User-Level Depression Detection}

While analyzing individual tweets gives us important clues about how text and images work together, a single post is rarely enough to understand a person's mental state. Depression is usually a pattern of behavior over time, not just a one-time event. Therefore, our main goal is to classify users based on their full history by looking at sequences of up to 512 tweets per user. 

In this section, we test three different methods to turn tweet-level predictions into a final classification for the user: Majority Voting, Softmax Probability Averaging, and a Sequential Transformer. For the methods that need a threshold which are Majority Voting and Softmax Averaging, we tested different probability cut-offs from 0.3 to 0.7. We found that a threshold of 0.5 always gave the best balance between precision and recall. Because of this, we use 0.5 as our standard decision point for all the results below.

\subsubsection{Majority Voting}

The simplest aggregation strategy we evaluated is Majority Voting. In this baseline, each tweet in a user's timeline is classified independently. A user is labeled as depressed if the proportion of their posts predicted as depressed exceeds a decision threshold $\theta$.

\begin{figure}[htbp]
\centering
\includegraphics[width=0.9\textwidth]{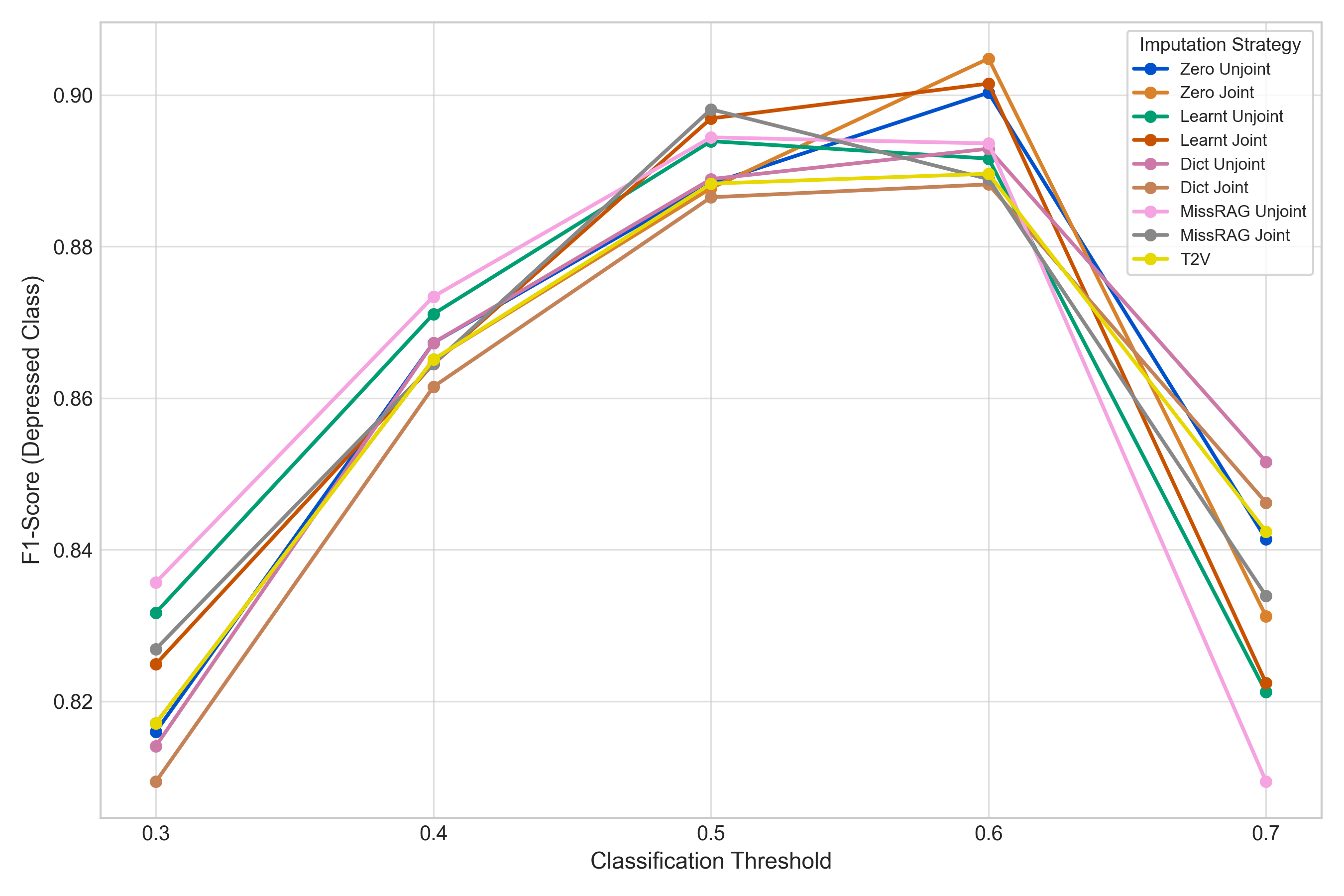}
\caption{Majority Voting: Depressed $F_1$-Score across various classification thresholds. Performance peaks at $\theta = 0.6$ across all imputation strategies.}
\label{fig:majority_thresholds}
\end{figure}

To find the best decision boundary, we tested threshold values from $0.3$ to $0.7$. As shown in Figure \ref{fig:majority_thresholds}, the $F_1$-score for the depressed class reaches its peak at $\theta = 0.6$ across all imputation strategies. Table \ref{tab:majority_voting} presents the user-level classification results (Depressed $F_1$-score and ROC-AUC) at this optimal threshold.

Aggregating tweet predictions over a user's entire timeline improves performance compared to isolated single-tweet classification. Setting the decision threshold to $0.6$ helps filter out false-positive predictions from individual posts, allowing all multimodal imputation methods to achieve $F_1$-scores around $0.89\text{--}0.90$ and ROC-AUC scores around $0.89\text{--}0.91$.

Under this configuration, the baseline Zero Placeholder method reaches an $F_1$-score of $0.9048$ (ROC-AUC: $0.9091$) in the joint setting and $0.9003$ (ROC-AUC: $0.9062$) in the unjoint setting. The Constant Learnt baseline obtains $0.9015$ (ROC-AUC: $0.9063$) in the joint setting and $0.8916$ (ROC-AUC: $0.8977$) in the unjoint setting. The context-aware methods show similar stability: MissRAG reaches $0.8936$ (ROC-AUC: $0.9006$) in unjoint training and $0.8889$ (ROC-AUC: $0.8949$) in joint training, while the neural T2V network achieves $0.8896$ (ROC-AUC: $0.8949$). The proposed Vis-Dict method achieves an $F_1$-score of $0.8929$ (ROC-AUC: $0.8977$) in the unjoint setting and $0.8882$ (ROC-AUC: $0.8920$) in the joint setting. These results confirm that simple voting aggregation provides a solid, consistent baseline across all visual imputation strategies.

\begin{table}[htbp]
\centering
\caption{User-level classification results using Majority Voting (Threshold $\theta = 0.6$).}
\label{tab:majority_voting}
\small
\begin{tabular}{llcc}
\hline
\textbf{Imputation Strategy} & \textbf{Training Paradigm} & \textbf{Depressed $F_1$} & \textbf{ROC-AUC} \\
\hline
Zero Placeholder & Unjoint & 0.9003 & 0.9062 \\
                 & Joint   & 0.9048 & 0.9091 \\
\hline
Constant Learnt  & Unjoint & 0.8916 & 0.8977 \\
                 & Joint   & 0.9015 & 0.9063 \\
\hline
MissRAG          & Unjoint & 0.8936 & 0.9006 \\
                 & Joint   & 0.8889 & 0.8949 \\
\hline
T2V Network      & Joint   & 0.8896 & 0.8949 \\
\hline
Vis-Dict (Ours)  & Unjoint & 0.8929 & 0.8977 \\
                 & Joint   & 0.8882 & 0.8920 \\
\hline
\end{tabular}
\end{table}

\subsubsection{Softmax Probability Averaging}

While Majority Voting makes a hard binary decision for every post, Softmax Probability Averaging incorporates the continuous confidence scores from the classification head. In this method, we average the predicted probabilities for the depressed class across all posts in a user's timeline. If the mean probability exceeds a decision threshold $\theta$, the user is classified as depressed.

\begin{figure}[htbp]
\centering
\includegraphics[width=0.9\textwidth]{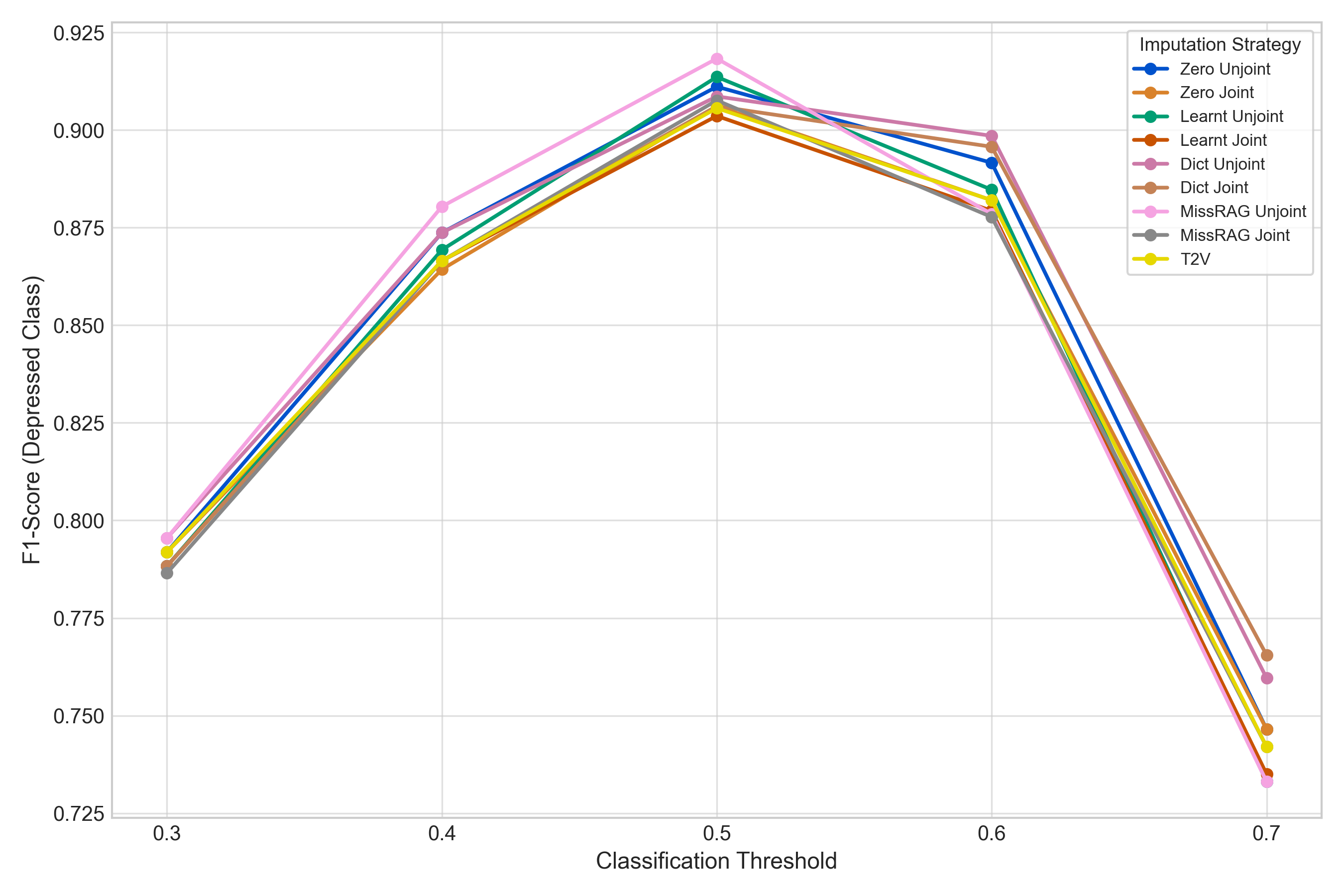}
\caption{Softmax Averaging: Depressed $F_1$-Score across various classification thresholds. Performance peaks at $\theta = 0.5$ across all imputation strategies.}
\label{fig:softmax_thresholds}
\end{figure}

We evaluated decision thresholds from $0.3$ to $0.7$. As shown in Figure \ref{fig:softmax_thresholds}, the Depressed $F_1$-score reaches its maximum at $\theta = 0.5$ across all imputation strategies. Table \ref{tab:softmax_averaging} presents the user-level classification results (Depressed $F_1$-score and ROC-AUC) at this optimal threshold.

Averaging continuous probabilities yields higher and more consistent scores across all configurations compared to hard voting, raising $F_1$-scores to $0.90\text{--}0.92$ and ROC-AUC scores to $0.90\text{--}0.92$. Retrieval-based MissRAG achieves an $F_1$-score of $0.9183$ (ROC-AUC: $0.9176$) in the unjoint setting and $0.9076$ (ROC-AUC: $0.9062$) in the joint setting. The Constant Learnt baseline obtains $0.9136$ (ROC-AUC: $0.9119$) in unjoint training and $0.9036$ (ROC-AUC: $0.9006$) in joint training, while the Zero Placeholder reaches $0.9111$ (ROC-AUC: $0.9091$) and $0.9061$ (ROC-AUC: $0.9034$). 

The proposed Vis-Dict method achieves an $F_1$-score of $0.9086$ (ROC-AUC: $0.9062$) in the unjoint setting and $0.9061$ (ROC-AUC: $0.9034$) in the joint setting, performing competitively with the neural T2V network ($0.9056$ / $0.9034$). These results indicate that aggregating continuous confidence values effectively preserves nuanced signals across the timeline prior to the final user-level decision.

\begin{table}[htbp]
\centering
\caption{User-level classification results using Softmax Probability Averaging (Threshold $\theta = 0.5$).}
\label{tab:softmax_averaging}
\small
\begin{tabular}{llcc}
\hline
\textbf{Imputation Strategy} & \textbf{Training Paradigm} & \textbf{Depressed $F_1$} & \textbf{ROC-AUC} \\
\hline
Zero Placeholder & Unjoint & 0.9111 & 0.9091 \\
                 & Joint   & 0.9061 & 0.9034 \\
\hline
Constant Learnt  & Unjoint & 0.9136 & 0.9119 \\
                 & Joint   & 0.9036 & 0.9006 \\
\hline
MissRAG          & Unjoint & 0.9183 & 0.9176 \\
                 & Joint   & 0.9076 & 0.9062 \\
\hline
T2V Network      & Joint   & 0.9056 & 0.9034 \\
\hline
Vis-Dict (Ours)  & Unjoint & 0.9086 & 0.9062 \\
                 & Joint   & 0.9061 & 0.9034 \\
\hline
\end{tabular}
\end{table}

\subsubsection{Sequential Transformer}

Majority Voting and Softmax Averaging are effective, but they treat a user's timeline as an unordered collection of posts. Because depression symptoms develop over time, the chronological sequence of posts provides important behavioral clues. To capture these long-term patterns, we evaluated a Sequential Transformer that processes timelines of up to 512 tweets per user in their exact posting order. To ensure reliability, all sequential experiments were run 5 independent times using different random seeds, and we report the Mean $\pm$ Standard Deviation.

Table \ref{tab:transformer_results} presents the user-level classification results across three settings: unimodal baselines, multimodal models, and single-image models using imputed data. The sequential model achieves the strongest overall performance in our study. The unimodal text baseline achieves a high $F_1$-score of $0.9359 \pm 0.0127$ and an ROC-AUC of $0.9869 \pm 0.0005$, confirming that linguistic patterns over time are a very strong indicator of depression.

In the multimodal setting, the proposed Vis-Dict method (Unjoint) delivers strong performance, achieving an ROC-AUC of $0.9890 \pm 0.0009$ and an $F_1$-score of $0.9454 \pm 0.0128$. Rather than showing outright superiority, this represents a statistical tie in $F_1$-score with the T2V network ($F_1 = 0.9468 \pm 0.0048$, $\text{ROC-AUC} = 0.9855 \pm 0.0008$) and performs comparably to the retrieval-based MissRAG ($F_1 = 0.9416 \pm 0.0097$, $\text{ROC-AUC} = 0.9869 \pm 0.0007$). Furthermore, Vis-Dict achieves this competitive discrimination power using zero trainable imputation parameters. This shows that the visual dictionary provides effective complementary information without needing extra training parameters for image generation. Additionally, when evaluating the models using only the filled-in visual features (Section 3 of Table \ref{tab:transformer_results}), Vis-Dict achieves an $F_1$-score of $0.8821 \pm 0.0136$, showing that these estimated features still carry useful information on their own.

\begin{table}[htbp]
\centering
\caption{User-level classification performance using the Sequential Transformer across 5 runs.}
\label{tab:transformer_results}
\small
\begin{tabular}{llcc}
\hline
\textbf{Model Configuration} & \textbf{Training Paradigm} & \textbf{Depressed $F_1$} & \textbf{ROC-AUC} \\
\hline

\multicolumn{4}{l}{\textbf{1. Unimodal Baselines}} \\
\hline
Text-Only (All Samples)          & Frozen  & $0.9359 \pm 0.0127$ & $0.9869 \pm 0.0005$ \\
Text-Only (Full Modality)       & Frozen  & $0.8915 \pm 0.0023$ & $0.9683 \pm 0.0007$ \\
Image-Only (Full Modality)      & Unjoint & $0.8611 \pm 0.0234$ & $0.9361 \pm 0.0048$ \\
Image-Only (Full Modality      & Joint   & $0.7747 \pm 0.0506$ & $0.8707 \pm 0.0041$ \\
\hline

\multicolumn{4}{l}{\textbf{2. Multimodal (w/ Imputation)}} \\
\hline
Zero Placeholder            & Unjoint & $0.9237 \pm 0.0197$ & $0.9861 \pm 0.0022$ \\
Zero Placeholder            & Joint   & $0.9077 \pm 0.0131$ & $0.9804 \pm 0.0024$ \\
Constant Learnt             & Unjoint & $0.9215 \pm 0.0109$ & $0.9857 \pm 0.0012$ \\
Constant Learnt             & Joint   & $0.9021 \pm 0.0170$ & $0.9786 \pm 0.0043$ \\
MissRAG                     & Unjoint & $0.9416 \pm 0.0097$ & $0.9869 \pm 0.0007$ \\
MissRAG                     & Joint   & $0.9408 \pm 0.0075$ & $0.9855 \pm 0.0008$ \\
T2V Network                 & Joint   & $0.9468 \pm 0.0048$ & $0.9855 \pm 0.0008$ \\
Vis-Dict (Ours)             & Unjoint & $0.9454 \pm 0.0128$ & $0.9890 \pm 0.0009$ \\
Vis-Dict (Ours)             & Joint   & $0.9337 \pm 0.0084$ & $0.9855 \pm 0.0009$ \\
\hline

\multicolumn{4}{l}{\textbf{3. Image-Only (w/ Imputation)}} \\
\hline
MissRAG                     & Unjoint & $0.9180 \pm 0.0027$ & $0.9748 \pm 0.0007$ \\
MissRAG                     & Joint   & $0.9146 \pm 0.0109$ & $0.9773 \pm 0.0013$ \\
T2V Network                 & Joint   & $0.9155 \pm 0.0157$ & $0.9801 \pm 0.0016$ \\
Vis-Dict (Ours)             & Unjoint & $0.8821 \pm 0.0136$ & $0.9534 \pm 0.0022$ \\
Vis-Dict (Ours)             & Joint   & $0.8189 \pm 0.0197$ & $0.9167 \pm 0.0038$ \\
\hline
\end{tabular}
\end{table}

\noindent\textbf{Ablation Study on Positional Encoding}\\
To test how much the chronological order of tweets matters, we conducted an ablation study. We removed the positional encoding vectors from the Transformer, forcing the model to process each user's tweets in a random order as an unstructured set. We tested this on the T2V Network, MissRAG (Joint), and Vis-Dict (Unjoint) across 5 runs. The results are shown in Table \ref{tab:ablation_positional}.

\begin{table}[htbp]
\centering
\caption{Sequential Transformer performance with and without Positional Encoding (5 runs).}
\label{tab:ablation_positional}
\small
\begin{tabular}{llcc}
\hline
\textbf{Imputation Strategy} & \textbf{Positional Encoding} & \textbf{Depressed $F_1$} & \textbf{ROC-AUC} \\
\hline
Vis-Dict (Unjoint) & Included & $0.9454 \pm 0.0128$ & $0.9890 \pm 0.0009$ \\
Vis-Dict (Unjoint) & Removed  & $0.9416 \pm 0.0102$ & $0.9897 \pm 0.0010$ \\
\hline
T2V Network (Joint)& Included & $0.9468 \pm 0.0048$ & $0.9855 \pm 0.0008$ \\
T2V Network (Joint)& Removed  & $0.9024 \pm 0.0229$ & $0.9762 \pm 0.0057$ \\
\hline
MissRAG (Joint)    & Included & $0.9408 \pm 0.0075$ & $0.9855 \pm 0.0008$ \\
MissRAG (Joint)    & Removed  & $0.9249 \pm 0.0080$ & $0.9825 \pm 0.0011$ \\
\hline
\end{tabular}
\end{table}

Removing the positional encoding caused a significant performance drop for both the generative T2V network (from 0.9468 to 0.9024) and the retrieval-based MissRAG (from 0.9408 to 0.9249). This shows that these methods rely heavily on the specific time sequence of posts to fuse information effectively. In contrast, our Vis-Dict method remained highly stable even without time information ($F_1 = 0.9416$, ROC-AUC $= 0.9897$). Because the dictionary method accumulates vocabulary over the entire timeline, it provides a robust diagnostic signal regardless of the exact order of the posts.

\noindent\textbf{Performance Breakdown by Data Subset}\\
To better understand how the models handle missing data, we evaluated the multimodal Sequential Transformer on two distinct subsets of the test data:
\begin{enumerate}
    \item \textbf{Full Modality:} Users whose evaluated posts all contain genuine text and images (no imputation needed).
    \item \textbf{Missing Modality:} Users whose posts are missing images, requiring the models to use imputed visual vectors.
\end{enumerate}

Table \ref{tab:full_vs_missing_user} shows the performance on these specific subsets across 5 runs.

\begin{table}[htbp]
\centering
\caption{Sequential model performance separated by Full Modality and Missing Modality subsets (5 runs).}
\label{tab:full_vs_missing_user}
\small
\begin{tabular}{llcc}
\hline
\textbf{Imputation Strategy} & \textbf{Training Paradigm} & \textbf{Depressed $F_1$} & \textbf{ROC-AUC} \\
\hline
\multicolumn{4}{c}{\textbf{1. Full Modality (Real Images)}} \\
\hline
Real Image (Oracle) & Unjoint & $0.9062 \pm 0.0014$ & $0.9698 \pm 0.0022$ \\
Real Image (Oracle) & Joint   & $0.8684 \pm 0.0104$ & $0.9475 \pm 0.0050$ \\
\hline
\multicolumn{4}{c}{\textbf{2. Missing Modality (Imputed)}} \\
\hline
MissRAG             & Unjoint & $0.9423 \pm 0.0062$ & $0.9841 \pm 0.0002$ \\
MissRAG             & Joint   & $0.9425 \pm 0.0057$ & $0.9836 \pm 0.0007$ \\
T2V Network         & Joint   & $0.9425 \pm 0.0044$ & $0.9834 \pm 0.0006$ \\
Vis-Dict (Ours)     & Unjoint & $0.9434 \pm 0.0074$ & $0.9839 \pm 0.0005$ \\
Vis-Dict (Ours)     & Joint   & $0.9416 \pm 0.0045$ & $0.9838 \pm 0.0006$ \\
\hline
\end{tabular}
\end{table}

The breakdown reveals two important points. First, on the full modality subset, the unjoint training method ($0.9062$) performs noticeably better than joint training ($0.8684$), suggesting that keeping the visual features independent preserves distinguishing information better when real images are available. Second, on the missing modality subset, the proposed Vis-Dict (Unjoint) achieves the highest precision among all methods, recording an $F_1$-score of $0.9434$ and an ROC-AUC of $0.9839$.

\subsection{Discussion}

Our results highlight several important points from our study. First, the single-tweet tests show that text is the main signal for detecting depression in isolated posts, while images provide helpful extra context.

Second, our error analysis shows that filling in missing images is useful. In 1,467 cases, the estimated image features corrected mistakes made by the text model, showing that visual information can help when the text is unclear.

Finally, the user-level experiments show the benefit of tracking behavior over time across a user's whole timeline. Using the Sequential Transformer, all missing-image methods achieve strong results. Vis-Dict achieves an $F_1$-score of $0.9454 \pm 0.0128$ and an ROC-AUC of $0.9890 \pm 0.0009$, which is a statistical tie with the T2V network ($F_1 = 0.9468 \pm 0.0048$). Furthermore, Vis-Dict reaches this competitive performance without needing any trainable parameters for image generation, showing that a visual dictionary is an effective and practical solution for long-term depression detection.

\section{Conclusion}

This paper addressed the problem of missing images in multimodal depression detection on social media. We evaluated five ways to handle missing images, from simple placeholders to generative networks, and proposed a TF-IDF Visual Dictionary (Vis-Dict). We tested these methods at both the single-tweet and user levels, using a Sequential Transformer to track changes in behavior over time across user timelines.

Our results show that while text is the main sign of depression in a single post, combining text and image information over a user's full timeline is essential for accurate classification. In our sequential experiments, the proposed Vis-Dict method (Unjoint) achieved strong performance, reaching an ROC-AUC of 0.9890 and an $F_1$-score of 0.9454, which is a statistical tie with the generative T2V network. Most importantly, Vis-Dict achieves this competitive performance with zero trainable parameters for image generation. Furthermore, ablation tests showed that the dictionary features remained very stable even when the time order of posts was removed. These findings show that a visual dictionary is an effective and practical solution for handling missing images in long-term depression detection.

A limitation of Vis-Dict is that it depends on word-image pairs, so it can struggle to create useful visual features for short posts that lack clear keywords. Future work will include testing the method under different amounts of missing data, using optical character recognition (OCR) to read text inside images, and extending the approach to handle missing text. We also plan to test better fusion methods and explore large vision-language models to improve multimodal alignment.

\end{document}